\documentclass[10pt]{article} 
\usepackage[preprint]{rlc2024}

\usepackage{amssymb}            
\usepackage{mathtools}          
\usepackage{mathrsfs}           
\usepackage{graphicx}           
\usepackage{subcaption}         
\usepackage[space]{grffile}     
\usepackage{url}                

\usepackage{algorithm}
\usepackage{algpseudocode}
\usepackage{amsthm}
\usepackage{makecell}
\usepackage{wrapfig}

\newcommand{\Ac}{\mathcal{A}}
\newcommand{\Cc}{\mathcal{C}}

\newcommand{\Mc}{\mathcal{M}}

\newcommand{\Rc}{\mathcal{R}}
\newcommand{\Sc}{\mathcal{S}}
\newcommand{\Pc}{\mathcal{P}}

\DeclareMathOperator*{\argmax}{arg\,max}

\newtheorem{definition}{Definition}

\title{Safety Modulation: Enhancing Safety in Reinforcement Learning through Cost-Modulated Rewards}

\author{Hanping Zhang,\quad Yuhong Guo \\
    jagzhang@cmail.carleton.ca, yuhong.guo@carleton.ca\\
    School of Computer Science, 
    Carleton University, Canada 
}

\begin{document}

\maketitle

\begin{abstract}
Safe Reinforcement Learning (Safe RL) aims to train an RL agent to maximize its performance in real-world environments 
	while adhering to safety constraints, as exceeding safety violation limits can result in severe consequences. 
In this paper, we propose a novel safe RL approach called Safety Modulated Policy Optimization (SMPO),
which enables safe policy function learning 
within the standard policy optimization framework through safety modulated rewards. 
In particular, we consider safety violation costs as feedback from the RL environments that are parallel to the standard awards, 
and introduce a Q-cost function as safety critic to estimate expected future cumulative costs.   
	Then we propose to modulate the rewards using a cost-aware weighting function,
	which is carefully designed to ensure the safety limits 
	based on the estimation of the safety critic,
	while maximizing the expected rewards. 
	The policy function and the safety critic are simultaneously learned 
	through gradient descent during online interactions with the environment. 
	We conduct experiments using multiple RL environments and 
        the experimental results demonstrate that our method outperforms 
	several classic and state-of-the-art comparison methods 
	in terms of overall safe RL performance.
\end{abstract}

\section{Introduction}
Reinforcement Learning (RL) offers a robust learning framework for decision-making systems 
in real-world scenarios, particularly in fields like robotics, where agents learn optimal policies by interacting with their environments. In recent years, RL 
has gained further attention in several emerging domains, including video games \citep{vinyals2017starcraft,berner2019dota}, healthcare \citep{yu2021reinforcement}, and even fine-tuning Large Language Models (LLMs) \citep{ouyang2022training}.
As RL finds broader application, the importance of safety considerations 
has become increasingly evident. This is especially critical in domains involving human-related tasks with a low tolerance for safety violations, such as helicopter manipulation \citep{martin2009learning,koppejan2011neuroevolutionary} and autonomous vehicles \citep{wen2020safe}, 
and even in mitigating potential harms from sensitive information generated by LLMs \citep{ouyang2022training}.

The growing importance of safety considerations in RL 
has promoted the development of safe Reinforcement Learning (safe RL), 
aimed at maximizing the performance of RL agents while adhering to specific safety constraints
such as a maximum number of safety violations \citep{altman1999constrained,mihatsch2002risk,ray2019benchmarking}. 
The problem is often formulated as a Constrained Markov Decision Process (CMDP) \citep{altman1999constrained}. 
With the emergence of research in RL focused on modifying the learning objective and corresponding update rules \citep{schulman2015trust,schulman2017proximal}, recent studies have addressed safe RL by incorporating safety constraints into classic RL frameworks \citep{achiam2017constrained,yang2022cup}. Reward shaping \citep{tessler2018reward,ray2019benchmarking} is a commonly used technique in safe RL to solve the constrained optimization problem in CMDP. It modifies the policy learning process by incorporating a penalization term (often a Lagrangian multiplier) into the reward function based on the safety constraints. 
However, 
conventional reward shaping methods 
penalize the reward function solely based on the current cumulative cost, ignoring potential safety violations in future exploration. 
Consequently, they underestimate the near future safety violations, 
often exceeding the cost limit 
by encountering additional safety violations.  
Moreover, previous reward shaping methods are typically limited to penalizing the reward function 
using an additional penalization term 
and treating the penalized reward as a non-differentiable term with respect to the policy,
which hinders the efficiency of simultaneous policy update and safety enhancement.

In this paper, we introduce a novel approach called Safety Modulated Policy Optimization (SMPO), 
aimed at effectively learning a safe policy within standard RL frameworks. 
Addressing the limitations of conventional reward shaping methods, which overlook expected safety violations in the future 
and struggle to efficiently update safety considerations with the policy, 
we shift our focus to the expected cumulative cost throughout exploration instead of the current collection. 
Specifically, we consider safety violation costs and standard awards as dual feedbacks from the RL environments 
and introduce a Q-cost function as safety critic to estimate the expected future cumulative costs.   
Building upon this concept, we introduce a novel safety modulated reward mechanism
that modulates the rewards using a cost-aware weighting function, 
which is carefully designed to 
realize an effective resolution of CMDPs in an unconstrained form
by maximizing the expected modulated rewards.  
The SMPO algorithm learns a safe policy through policy gradient methods 
by deploying the cost-aware weighted reward as a partially differentiable function of the policy,
which not only accelerates policy learning but also enhances safety. 
Experimental results validate that our proposed SMPO method can firmly adhere to safety violation constraints 
by maintaining the cumulative cost limit during exploration. 
Overall, it outperforms several classic and state-of-the-art safe RL methods in terms of performance-safety trade-off. 
%

\section{Related Works}
Safe RL addresses the challenge of efficiently learning a policy within an environment while satisfying specific safety constraints. 
Early research by \citet{altman1999constrained} 
has formally defined the constrained optimization problem essential to safe RL 
as a Constrained Markov Decision Process (CMDP). 
\citet{mihatsch2002risk} introduced the notion of risk in safe RL, 
focusing on learning risk-averse policies via risk-sensitive control processes. 
A comprehensive survey on safe RL has been conducted by 
\citet{garcia2015comprehensive}, 
outlining different research fields within the domain. 
More recently, \citet{ray2019benchmarking} 
provided an overview of standard safe RL problems and classic safe RL algorithms, 
offering an extensive testbed for advancing safe RL research.

\paragraph{Safe RL for Solving CMDPs}
The base challenge in safe RL lies in addressing CMDPs, which presents constrained optimization problems rather than the standard optimization framework in RL. This unique setting has prompted the development of various approaches aimed at tackling CMDPs in diverse ways.
Previous research in standard RL \citep{schulman2015trust} formulated the policy function update as a trust region optimization, introducing the Trust Region Policy Optimization (TRPO) method
to enhance the robustness of policy learning. 
Similarly, \citet{schulman2017proximal} proposed the Proximal Policy Optimization (PPO) algorithm, 
which utilizes a clipped surrogate objective function 
to facilitate stable and efficient policy updates.
Building upon such prior research in standard policy gradient methods, 
\citet{achiam2017constrained} introduced the Constrained Policy Optimization (CPO) algorithm, which constrains the difference between the rewards or costs of two different policies to update the safe policy by optimizing the primal-dual problem within trust regions.
In a more recent work, \citet{yang2022cup} introduced a conservative update policy (CUP) 
algorithm, utilizing tight performance bounds as surrogate functions to design safe RL algorithms. 
\citet{xu2022constraints} proposed the Constraint Penalized Q-learning (CPQ) method, employing a constraint penalized bellman operator to stop the update of the Q function when the safe RL agent enters potentially unsafe states.
Additionally, the Lagrangian method, a popular constrained optimization technique, can transform a CMDP problem into 
an unconstrained 
optimization problem by incorporating safety constraints 
as penalization terms in the objective 
through the introduction of Lagrange multipliers. 
\citet{ray2019benchmarking} provided a comprehensive introduction to such methods and offer them as benchmarks for safe RL. 
\citet{tessler2018reward} further
introduced the reward shaping technique to the standard actor-critic algorithm, leading to the Reward Constrained Policy Optimization (RCPO), which learns the policy through an alternative penalized reward signal. However, previous reward shaping algorithms primarily focus on penalizing the reward function based on current satisfaction with the safety constraints, without considering expected 
safety violations in future exploration. 
Previous works also treat the penalized reward as a non-differentiable term with respect to the policy, 
which hinders the efficient updating of the policy function.
\paragraph{Safe RL with Safety Critics}
To address the challenge of avoiding potential safety violations in exploration, 
several recent studies have explored 
the application of safety critics in safe RL. 
A safety critic is trained in the same way as the standard critic function 
and serves as guidance to avoid potential unsafe states during exploration.
\citet{srinivasan2020learning} 
introduced the approach of Safety Q-functions for Reinforcement Learning (SQRL), 
which learns a safety critic to estimate the probability of future safety violations, enabling the avoidance of potentially unsafe states by stopping policy updates when the critic's prediction exceeds a threshold. 
Expanding on this work, \citet{bharadhwaj2020conservative} adopted Conservative Q-Learning (CQL) \citep{kumar2020conservative} 
to develop a safety critic that overestimates the likelihood of catastrophic failures, enhancing safety measures. 
Their approach involves iteratively resampling actions until a sufficiently safe action is identified when the safety critic's prediction falls below a specified threshold. 
However, these approaches adopt a different problem setting where no safety violations are permitted during the exploration process. 
In these methodologies, the safety critic serves as a predictor of unsafe states and is utilized directly to avoid potentially risky exploration. 
By contrast, 
our approach diverges by training a safety critic to estimate the future cumulative cost, 
supporting a strategically designed reward weighting function that ensures the safety constraints in CMDPs.

\section{Problem Setting}
The safe RL problem is typically framed as a Constrained Markov Decision Process (CMDP)~\citep{altman1999constrained}, denoted as 
$M=(\Sc, \Ac, \Pc, \Rc, \Cc, \gamma)$, where $\Sc$ represents the state space, $\Ac$ is the action space, 
$\Pc: \Sc\times \Ac\to \Sc$ defines the transition probabilities,
$\Rc: \Sc\times \Ac\to \mathbb{R}$ is the reward function, 
and $\gamma\in (0, 1)$ is the discount factor. 
In addition to the reward function, a cost function $\Cc: \Sc\times \Ac\to [0, 1]$ is introduced to account for safety violations during RL exploration,
where 
the cost is 1 when a safety violation occurs. 
An exploration trajectory within CMDP can be expressed as 
$\tau=(s_0, a_0, r_0, c_0, \ldots, s_t, a_t, r_t, c_t, \ldots)$. 
The objective of CMDP is to learn a policy $\pi_\theta$ to maximize the expected discounted cumulative reward,  
denoted as $J^r_\pi(\theta)$, while ensuring adherence to safety constraints, 
where the cumulative cost $J^c_\pi(\theta)$ must not exceed a threshold $d$ that represents 
the maximum tolerance for safety violations: 
\begin{align}
\label{eqa:cmdp}
	\pi_{\theta^\star}=\argmax_{\pi_\theta} \; J^r_\pi(\theta)&=\mathbb{E}_{\tau\sim \pi_\theta}\left[\sum\nolimits_{t=0}^{\infty} \gamma^t r_t\right] 
	\quad 
	\quad \text{s.t. } J^c_\pi(\theta)=\mathbb{E}_{\tau\sim \pi_\theta}\left[\sum\nolimits_{t=0}^{\infty} \gamma^t c_t\right] \leq d
\end{align}
%

\section{Method}

\begin{figure*}[t]
\centering
\includegraphics[width=.9\textwidth]{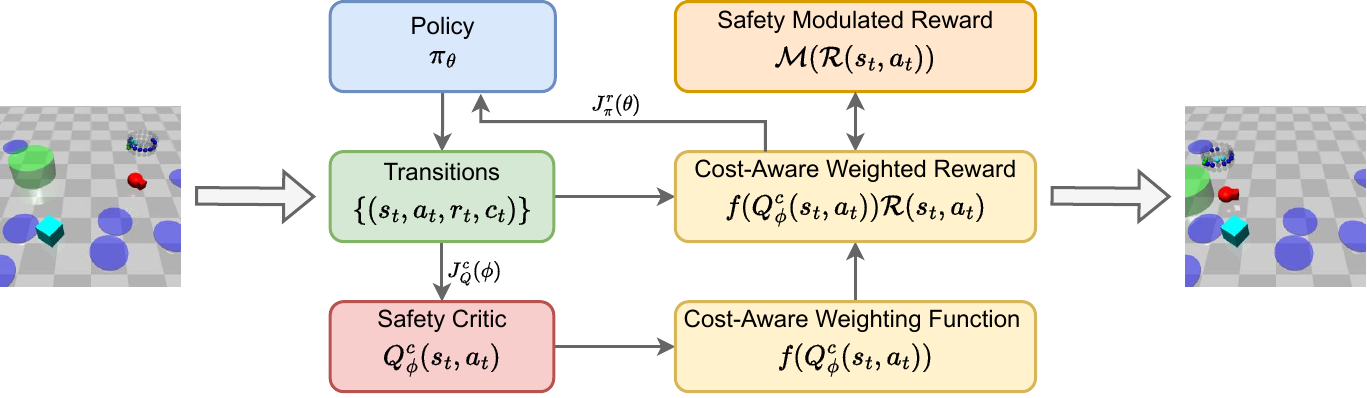}
\caption{
The main framework of our proposed SMPO approach. 
	At each timestep $t$, transitions $\{(s_t,a_t, r_t,c_t)\}$ are collected through exploration employing the policy $\pi_\theta$. 
	The observed cost $c_t$ contributes to the training of the safety critic $Q_\phi^{c}(s_t,a_t)$. 
	Utilizing the safety critic, the cost-aware weighting function $f(Q_\phi^{c}(s_t,a_t))$ is applied to 
	modulate the reward in a bilinear form
	$f(Q_\phi^{c}(s_t,a_t))\Rc(s_t,a_t)$ 
	to facilitate safe policy learning. 
}
\label{fig:model}
\end{figure*}
%
In this section, we present the proposed Safety Modulated Policy Optimization (SMPO) method for safe RL. 
The core idea is centered on the introduction of a safety modulation concept and the 
design of a strategic reward weighting function based on the expected cumulative cost of the exploration,
which encodes the safety constraints of CMDPs under the expected cumulative reward maximization framework of standard RL. 
Concretely, 
we learn a safety critic to estimate the expected future cumulative cost in explorations starting from the current state.  
Based on this estimation, we introduce the concept of safety modulation by 
defining a reward modulation function that reshapes the reward to encode safety constraints.
To conveniently utilize this function for safe policy optimization, we subsequently approximate
the conceptual reward modulation function by designing a cost-aware weighting function 
to modulate the reward in a bilinear form. 
The main framework of our proposed method is presented in Figure \ref{fig:model}.

\subsection{Learning Safety Critics}
To ensure the safety constraints in safe RL defined as CMDPs,
in addition to the current cumulative cost, we also need to consider the cumulative discounted cost for 
potential future explorations starting from the current state. 
To this end, 
inspired by Q-value function, $Q(s_t, a_t)$, 
that estimates the future cumulative discounted reward for a given state-action pair $(s_t, a_t)$, 
we consider safety violation costs and standard rewards as parallel feedbacks from the RL environment, 
and propose to learn a similar Q-cost function, $Q^c(s_{t}, a_{t})$,
to estimate the future cumulative discounted cost starting from $(s_t, a_t)$. 
This Q-cost function can serve as a safety critic for learning the safe policy function. 
Henceforth, we will interchangeably use terms Q-cost function and safety critic .

Different from the Q-value function which estimates the rewards that are the maximization target of RL,  
the costs are only produced as parallel passive feedback of the environment from the reward triggered actions. 
Therefore, we are unable to select actions that maximize the Q-cost for the subsequent state
as in Q-learning \citep{sutton2018reinforcement}. 
Instead, 
we propose to compute the estimation of the Q-cost value for each state-action pair $(s_t, a_t)$,
$\hat{Q}_\phi^c(s_t,a_t)$, 
by using the policy function 
in a similar style as in the expected SARSA~\citep{sutton2018reinforcement} 
as follows:
\begin{align}
	\hat{Q}^c(s_t,a_t)=\mathcal{C}(s_t,a_t)+\gamma \mathbb{E}_{a_{t+1}\sim \pi_\theta(\cdot|s_{t+1})}Q_\phi^c(s_{t+1}, a_{t+1})
	\label{eq:Qestimate}	
\end{align}
To learn a Q-cost function, $Q_\phi^c$, parameterized with $\phi$, 
we minimize the Mean Squared Error (MSE) between the predictions from the Q-cost function and the 
estimated values $\hat{Q}^c(s_t,a_t)$. 
Moreover, to prevent overestimation of future risk, we include an $\ell_2$-norm regularization term for the Q-cost function
and formulate the following minimization objective for safety critic learning: 
\begin{align}
\label{eqa:safety_critic}
J_Q^c(\phi)=\mathbb{E}_{(s_t,a_t)\sim\pi_\theta}\left[
	(Q_\phi^c(s_t,a_t) - \hat{Q}^c(s_t,a_t))^2
+\lambda \left \| Q_\phi^c(s_t,a_t) \right \|_2^2\right]
\end{align}
where $\lambda$ is the trade-off hyperparameter. 
Utilizing the learned safety critic $Q_\phi^{c}(s_t,a_t)$, which estimates the future cumulative cost, 
we are able to accurately estimate the expected cumulative cost for the current exploration 
as $\sum_{k=0}^{t-1}c_k + Q_\phi^c(s_t,a_t)$.

\subsection{Safety Modulation}
Constrained optimization problems have long been challenging topics in optimization. 
Directly solving the safe RL problem formulated as a CMDP in Eq.(\ref{eqa:cmdp}) is difficult. 
Instead, we propose a novel safety modulation framework to 
transform the original constrained optimization problem of safe RL 
into a standard reinforcement learning problem through a safety modulated reward function. 
To prevent from visiting the unsafe states that lead to exceeding of the safe violation limit,
we present the following definition for our safety modulated reward function. 
\begin{definition}
\label{def:unsafe}
For a given state-action pair $(s_t, a_t)$, the state $s_t$ becomes unsafe if the expected cumulative cost 
$\sum_{k=0}^{t-1} c_k+Q^c_\phi(s_t, a_t)$ exceeds the cost threshold $d$ by taking action $a_t$. 
In response to this unsafe condition,
a safety modulated reward function $\Mc(\Rc(s_t, a_t))$ severely penalizes the reward 
by transforming it to a sufficiently small negative value $-N$ (e.g., $N=\infty$ in the extreme) 
such that selecting action $a_t$ at $s_t$ will conflict with the cumulative reward maximization principle of standard RL. 
Alternatively, if the expected cumulative cost does not exceed the threshold, 
the safety modulated reward function preserves the original reward. 
In summary, our safety modulated reward function is defined as the following piecewise function:	
\begin{align}
\label{eqa:penalized}
\Mc(\Rc(s_t,a_t))=\left\{
\begin{aligned}
&\Rc(s_t,a_t), &&\text{if} \quad \sum\nolimits_{k=0}^{t-1} c_k+Q_\phi^c(s_t, a_t) \leq d\\
&-N, &&\text{else}
\end{aligned}
\right.
\end{align}
\end{definition}
Based on this definition, 
with good estimations of expected future costs through the safety critic, 
we anticipate automatic adherence to the constraint on safety violation limits
by conducting standard unconstrained RL learning with the safety modulated rewards.  

Nevertheless, 
Definition \ref{def:unsafe} establishes the concept of safety modulation, 
but provides a non-differentiable 
reward modulation function $\Mc$,
which is a function of the parametric safety critic $Q_\phi^c$.
To address this problem and facilitate safe RL, 
we propose to realize $\Mc$ in the following
bilinear form by using a cost-aware weighting function $f(\cdot)$ to reshape the original reward: 
\begin{align}
\label{eqa:safety_modulated}
\Mc(\Rc(s_t,a_t))=f(Q_\phi^c(s_t,a_t))\cdot \Rc(s_t,a_t).
\end{align}
As such, we can design the weighting function 
$f(Q_\phi^c(s_t,a_t))$ to model the conditions in the piecewise function $\Mc$
and reshape the original reward accordingly.

\subsection{Cost-Aware Weighting Function}
The goal is to design a differentiable cost-aware weighting function $f(Q_\phi^c(s_t,a_t))$
such that the safety modulated reward function $\Mc$ 
in Definition \ref{def:unsafe} can be maximally approximated  
by weighting/scaling the original rewards through Eq.(\ref{eqa:safety_modulated}).  
To this end, we define the cost-aware weighting function
as an exponential function of the total estimated cost for the given state-action pair, 
i.e., $(\sum_{k=0}^{t-1}c_k+Q_\phi^c(s_t,a_t))$,
and hence a function of the safety critic,
as follows:
\begin{align}
f(Q_\phi^c(s_t,a_t))=\frac{b^d}{b^d-1}\left[1-b^{\big(\sum_{k=0}^{t-1}c_k+Q_\phi^c(s_t,a_t)\big)-d}\right]
	\label{eq:weightfunc}	
\end{align}
where the base $b$ is a constant that 
determines how the weighting function value changes in response to variations in the input total estimated cost. 
We assume $b>1$ to ensure that the weighting function is a decreasing function of the input total estimated cost. 
Importantly, when the total estimated cost is $0$, the weighting function returns a weight value $1$
to exactly represent the modulation function $\Mc$ and maintain the original reward through 
Eq.(\ref{eqa:safety_modulated});
when the total estimated cost is equal to the cost threshold $d$, 
the weighting function value becomes $0$,
and then quickly decreases in exponential rate to large negative values
to significantly penalize the reward 
when the total estimated cost further increases after exceeding the cost threshold $d$.

\begin{wrapfigure}{R}{0.45\textwidth}
\centering
\includegraphics[width=0.40\textwidth]{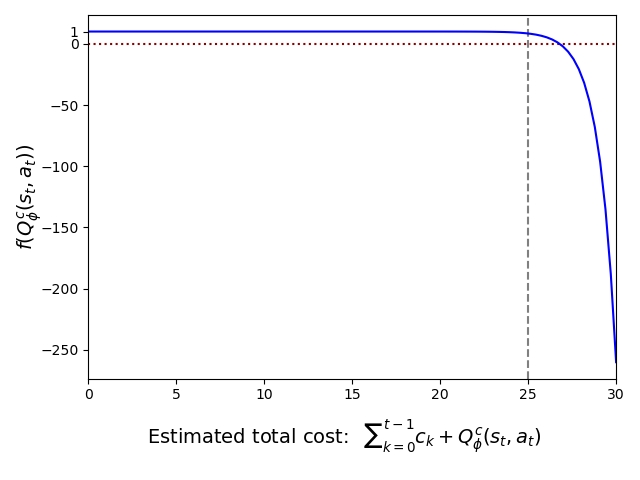}
	\vspace{-.05in}
\caption{
Visualization of the cost-aware weighting function $f(Q_\phi^{c}(s_t,a_t))$ with a base value $b=3$. 
We utilize a fixed safety cost threshold of $d=25$, indicated by the vertical dashed grey line. 
Additionally, the horizontal dotted line in dark red illustrates the scenario 
	where $f(Q_\phi^{c}(s_t,a_t))=0$ when the total estimated cost reaches the cost threshold $d$. 
}
\label{fig:function}
	\vspace{-.40in}
\end{wrapfigure}
%
In Figure \ref{fig:function}, we present a visualization example of 
our strategically designed cost-aware weighting function $f(\cdot)$ 
with a base value of $b=3$
and a cost threshold of $d=25$ indicated by the grey dashed line. 
The horizontal dotted line in dark red color illustrates the scenario 
where $f(Q_\phi^{c}(s_t,a_t))=0$, signifying that the reward is penalized to 0
when the total estimated cost is equal to the cost threshold $d$. 
Notably, the function curve in blue color is almost horizontal around value 1
before the total estimated cost increases from $0$ to a value that is 
relatively very close to the cost threshold $d$,
indicating that the weight function maximally preserves the original reward
by producing weight values close to 1.
After the total estimated cost exceeding the cost threshold $d$,
the function value 
quickly drops to large negative weight values,
which can effectively approximate the safety modulated 
reward function $\Mc$
when multiplying by the original positive reward. 
This example demonstrates that the proposed cost-aware weighting function
can nicely approximate the safety modulated reward function $\Mc$
defined in Definition \ref{def:unsafe}  
though Eq.(\ref{eqa:safety_modulated}).

In addition, as a smooth function of the input and hence of the safety-critic, 
the gradient of the 
designed weighting function with respect to the input safety critic $Q_\phi^c$ can be 
conveniently derived as follows: 
\begin{align}
\label{eqa:gradient_reward}
\nabla_{Q_\phi^c} f(Q_\phi^{c}(s_t,a_t))\!=\!\frac{-b^d\ln b}{b^d-1}
	b^{\big(\sum_{k=0}^{t-1}c_k+Q_\phi^c(s_t,a_t)\big)-d}
\end{align}
Practically, the safety critic $Q_\phi^{c}(s_t,a_t)$, responsible for predicting future cumulative costs, should output non-negative values since costs cannot be negative. 
Furthermore, while there is no pre-given upper bound for potential costs in the exploration process, 
it is preferable to avoid overestimation of the future cost. 
To facilitate exploration while maintaining safety constraints, 
we mandate that the estimation of the safety critic for the discounted cumulative future cost does not surpass the cost threshold $d$.
This upper limit also helps prevent potential overflow during the calculation of the weighting function. 
Consequently, the practically used safety critic value
for computing the weight 
is clipped into the interval $[0,d]$, 
such as
$\text{clip}(Q_\phi^{c}(s_t,a_t), 0, d)=\max(0, \min(d,Q_\phi^{c}(s_t,a_t))).$
%

\subsection{Policy Optimization with Safety Modulated Reward}

Given the safety modulated rewards, safe RL can be conducted within the standard policy optimization framework 
by maximizing an expected modulated reward function $J_\pi^r(\theta)$. 
Let $d(s)$ denote the state distribution, where an action $a$ is sampled from the policy function $\pi_\theta(s, a)$. 
The objective function $J_\pi^r(\theta)$ can then be written 
as the expected safety modulated reward $\Mc(\mathcal{R}(s,a))$ 
over a joint distribution of states and actions: 
\begin{align}
J_\pi^r(\theta)
	&=\mathbb{E}_{\pi_\theta}[\Mc(\mathcal{R})]=\sum\nolimits_{s\in \Sc}d(s)\sum\nolimits_{a\in \Ac}\pi_\theta(s,a)\Mc(\Rc(s,a))
\end{align}
This objective function can also be formulated as the expected cumulative discounted safety-modulated reward according to the policy $\pi_\theta$ \citep{sutton1999policy, sutton2018reinforcement}:
\begin{align}
\label{eq:Jrpi}	
J_\pi^r(\theta)
	&=\mathbb{E}_{\tau\sim\pi_\theta}\left[\sum\nolimits_{t=0}^T\gamma^t \Mc(\Rc(s_t,a_t))\right]
	=\mathbb{E}_{\tau\sim\pi_\theta}\left[\sum\nolimits_{t=0}^T\gamma^t f(Q_\phi^c(s_t,a_t))\Rc(s_t,a_t)\right]
\end{align}
As the safety critic, i.e., the Q-cost function $Q_\phi^c$, is computed based on the policy function $\pi_\theta$,
the cost-aware weighting term $f(Q_\phi^c(s_t,a_t))$ is naturally a function of the policy $\pi_\theta$, 
which makes the policy gradient of the objective above fundamentally different from the case in standard RL. 
With a differentiable weighting term, 
we expect the extra gradient term can enhance the policy update efficiently.  
Consequently, the policy gradient of $J_\pi^r(\theta)$ can be derived as follows:
\begin{align}
\label{eqa:gradient_policy}
	\nabla_\theta J_\pi^r(\theta)=\mathbb{E}_{\tau\sim\pi_\theta}\Big[
&\sum\nolimits_{t=0}^T\gamma^t f(Q_\phi^c(s_t,a_t))\Rc(s_t,a_t)\nabla_\theta\log\pi_\theta(s_t,a_t) \nonumber \\
	+ &\sum\nolimits_{t=0}^T\gamma^t \Rc(s_t,a_t)\nabla_{Q_\phi^c} f(Q_\phi^c(s_t,a_t))\nabla_\theta Q_\phi^c(s_t,a_t)\Big]
\end{align}
To calculate this gradient, we need to compute both $\nabla_{Q_\phi^c} f(Q_\phi^c(s_t,a_t))$
and $\nabla_\theta Q_\phi^c(s_t,a_t)$ in addition to the standard policy gradient term 
$\nabla_\theta\log\pi_\theta(s_t,a_t)$.   
The gradient of the cost-aware weighting function with respect to $Q^c_\phi$ has already been derived 
in Eq.(\ref{eqa:gradient_reward}). 
To derive the gradient of the safety critic with respect to the policy parameters, 
$\nabla_\theta Q_\phi^c(s_t,a_t)$, 
we formulate the Q-cost function using the following Bellman equation, 
mirroring the structure of the Bellman equation for the Q-value function: 
\begin{align}
Q_\phi^c(s_t,a_t)\coloneqq \Cc(s_t,a_t)
	+\gamma\sum\nolimits_{s\in\Sc}\Pc(s|s_t,a_t)\sum\nolimits_{a\in\Ac}\pi_\theta(a|s)Q_\phi^c(s,a)
\end{align}
For simplicity, we further deploy the empirical transition $(s_t, a_t, s_{t+1})$ in the trajectory 
to replace the transition distribution
and approximate the Bellman equation as follows:
\begin{align}
Q_\phi^c(s_t,a_t)\approx \Cc(s_t,a_t)+\gamma\sum\nolimits_{a\in\Ac}\pi_\theta(a|s_{t+1})Q_\phi^c(s_{t+1},a)
\end{align}
The gradient of the safety critic with respect to the policy parameters $\theta$ can then be computed as:
\begin{align}
\nabla_\theta Q_\phi^c(s_t,a_t)
	=&\gamma \sum\nolimits_{a\in\Ac}Q_\phi^c(s_{t+1},a)\nabla_\theta\pi_\theta(a|s_{t+1})\nonumber\\ 
	=&\gamma\, \mathbb{E}_{a\sim\pi_\theta(\cdot|s_{t+1})}\left[Q_\phi^c(s_{t+1},a)\nabla_\theta\log\pi_\theta(a|s_{t+1})\right]
\label{eqa:gradient_critic}
\end{align}

During the online safe reinforcement learning process,  
the policy function and the safety critic are simultaneously learned 
through gradient descent 
by maximizing $J^r_\pi(\theta)$ in Eq.(\ref{eq:Jrpi})
and minimizing $J^c_Q(\phi)$ in Eq.(\ref{eqa:safety_critic}), respectively,
with data collected from the environment.

\subsection{Dynamic Schedule of Cost Threshold}
While our goal is to learn an optimal policy with the cumulative cost $J^c_\pi(\theta)$ strictly below the cost threshold $d$, fixing the cost threshold $d$ during early training can lead to practical challenges. 
In the early stage, 
the safety critic $Q_\phi^{c}(s_t,a_t)$ may not accurately predict future cumulative costs, potentially resulting in over-penalization of rewards due to imprecise cost predictions. 
Additionally, we expect the agent to initially focus more on exploring environments to 
gain knowledge and improve general performance 
before gradually prioritizing safety. 
If the agent prematurely converges to a safe region without sufficient exploration, 
the policy function may not be adequately learned, resulting in suboptimal performance.  
To address these issues, we deploy a dynamic schedule for the cost threshold value during training. 
Specifically, 
the cost threshold is initially set to $d'=\eta\cdot d$ with $\eta > 1$.
Then it is updated in each epoch according to the following schedule: 
\begin{align}
	d'=\left(\eta-(\eta-1)\frac{\min(e_{\max},e_p)}{e_{\max}}\right)\cdot d
	\label{eq:dschedule}
\end{align}
where $e_{\max}\geq 1$ is a constant integer that controls the number of epochs with varying cost thresholds, 
and $e_p$ denotes the current epoch number. 
Following this schedule, 
after $e_{\max}$ epochs of reinforcement exploration, the cost threshold $d'$ will be reduced to the fixed value of the pre-given $d$.

\section{Experiment}
%
\begin{figure}[ht]
\centering
    {\includegraphics[width=0.32\textwidth]{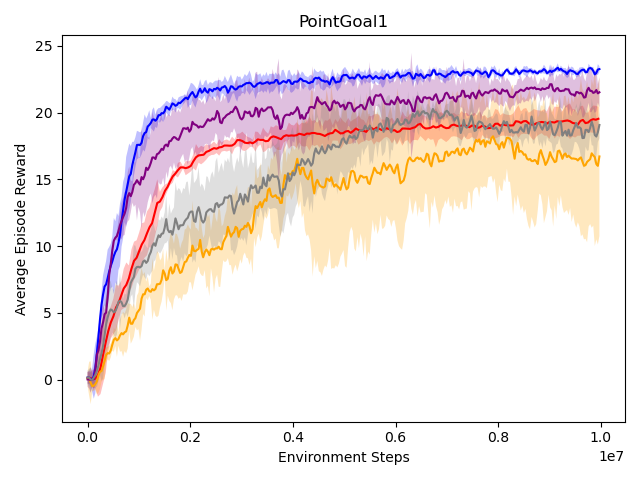}} 
	{\includegraphics[width=0.32\textwidth]{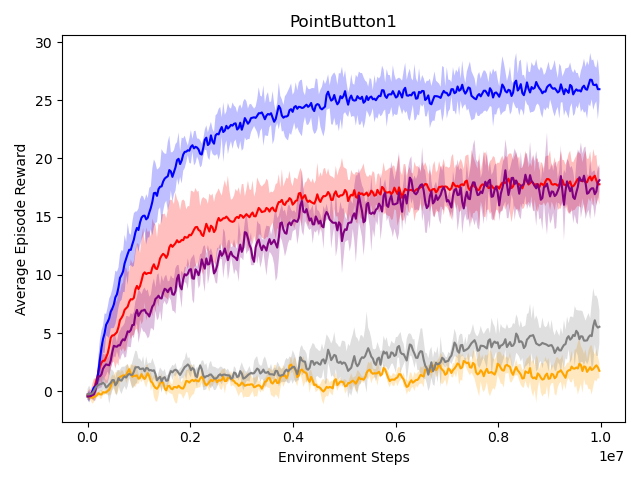}}
	{\includegraphics[width=0.32\textwidth]{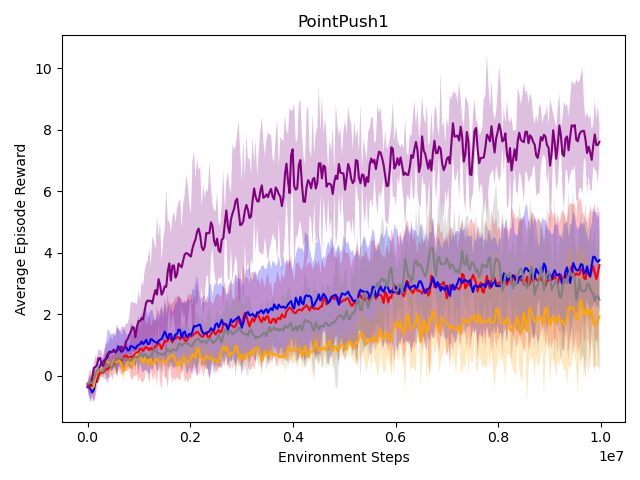}}
    {\includegraphics[width=0.32\textwidth]{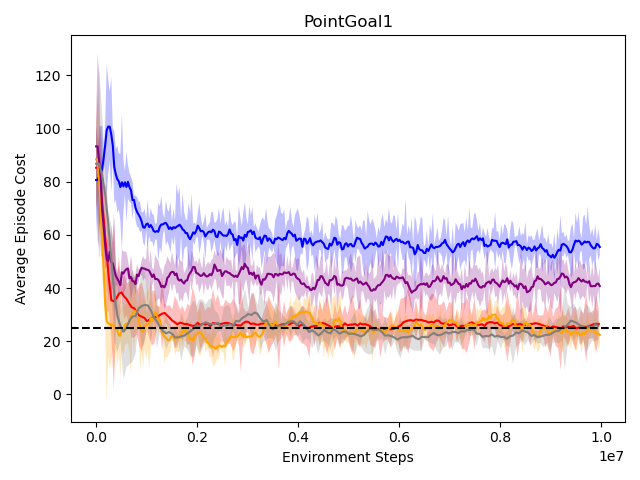}} 
	{\includegraphics[width=0.32\textwidth]{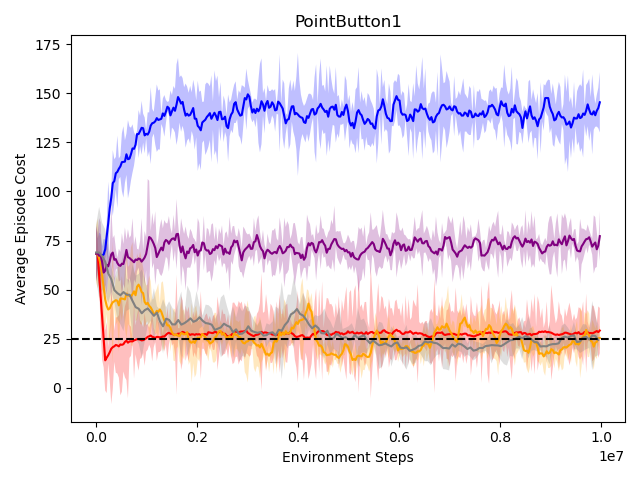}}
	{\includegraphics[width=0.32\textwidth]{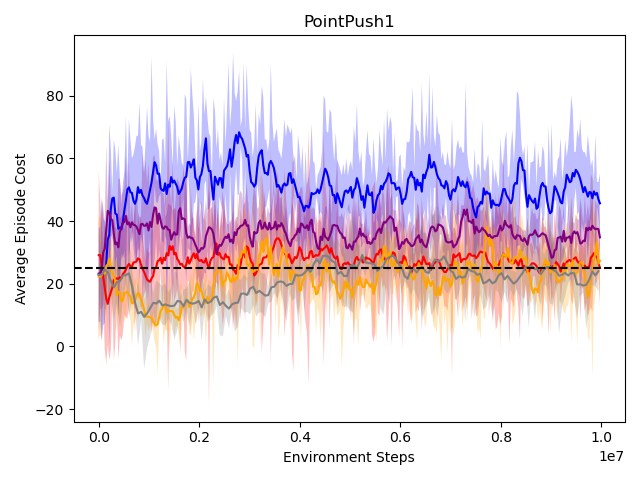}}
    {\includegraphics[width=0.5\textwidth]{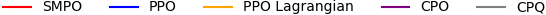}}
\caption{
Comparison results in terms of average episode reward/cost vs. environment steps on the Point robot across three different tasks: Goal1, Button1, and Push1. 
The top row reports the results in terms of average episode rewards,
while the bottom row reports the corresponding average episode costs. 
The results are averaged over three runs, with the shadow indicating standard deviations and the dashed black line indicating the cost threshold.
}
\label{fig:exp1}
\vskip -.15in
\end{figure}

\subsection{Experimental Settings}

\paragraph{Environments}
We conduct experiments using the Safety Gym platform \citep{ray2019benchmarking}, 
which facilitates research in safe RL through MuJoCo simulation tasks \citep{todorov2012mujoco}. 
The Safety Gym provides a comprehensive set of safe reinforcement learning tasks, including scenarios with multiple agents and varying difficulty levels. 
We performed experiments in six distinct environments: PointGoal1, PointButton1, PointPush1, PointGoal2, CarGoal1, and CarGoal2. 
In the PointGoal1, PointButton1, and PointPush1 environments, 
we evaluated the performance of a single robot agent (Point) across three respective tasks: 
reaching a designated goal, pressing a button at a specified location, and pushing an object to a target area. 
In the PointGoal2 environment, we evaluated the Point agent on a more challenging task Goal2 
with more hazards and obstacles. Furthermore, we investigated the performance of a Car agent in the CarGoal1 and CarGoal2 environments, both of which feature the Goal task at varying difficulty levels.
\paragraph{Comparison Methods}
We compared our proposed SMPO method with four methods: 
Proximal Policy Optimization (PPO) \citep{schulman2017proximal}, 
PPO Lagrangian \citep{ray2019benchmarking}, 
Constrained Policy Optimization (CPO) \citep{achiam2017constrained}, 
and Constrained Policy Q-Learning (CPQ) \citep{xu2022constraints}. 
The standard PPO method is deployed as a baseline that does not consider safety constraints. 
The other three comparison methods are specifically designed for safe RL. 
PPO Lagrangian is a safe RL variant of the PPO algorithm that transforms the constrained optimization problem 
into an unconstrained optimization problem using Lagrange multipliers. 
CPO is a safe RL method that takes costs into consideration and updates the safe policy by optimizing the primal-dual problem within trust regions. CPQ modifies the Bellman operator during the update of the Q function, stopping updates 
when the safe RL agent is in potentially unsafe situations.
\vskip -.25in
\paragraph{Implementation Details}
The proposed method builds primarily upon the PPO algorithm \citep{schulman2017proximal}, implemented within the Safety Gym environments \citep{ray2019benchmarking}. 
The safety critic is implemented as a 4-layer MLP with ReLU activation for the output layer. 
The policy function is implemented as a 4-layer MLP with tanh activation, 
employing softmax activation for discrete action spaces 
and Gaussian activation for continuous action spaces in the output layer.
Consistent with the recommendation by \citet{ray2019benchmarking}, 
we set the cost threshold $d$ equal to 25, the discount factor $\gamma$ equal to 0.99, 
and the maximum episode (i.e., trajectory) length $T$ equal to 1000.  
Each epoch consists of 30,000 environment steps.
For SMPO, we set $\eta=2$ and $e_{\max}=50$ for the dynamic schedule of cost threshold. 
We used $b=3$ for the cost-aware weighting function and set $\lambda=0.1$ for safety critic learning.
For the comparison methods, we utilize the official implementations from \citep{berner2019dota} 
for PPO, PPO Lagrangian, and CPO.
The implementation of CPQ is adopted from the offline safe RL repository 
\citep{liu2023constrained}, with modifications enabling online training. 
All experimental results are collected over a total of $10^7$ environmental steps.
%
\subsection{Experimental Results}
We employ the evaluation metrics outlined in \citep{ray2019benchmarking}, 
and report the averages of cumulative episodic reward and cost 
separately in each epoch.
Our objective is to optimize the RL agent's performance in terms of average cumulative episode reward 
while ensuring 
that the cumulative cost per episode does not exceed the specified cost limit.
\paragraph{Exploring Diverse Tasks}
We first conducted experiments 
for safe RL using 
diverse tasks with Point robots, 
evaluating the adaptability of safe RL methods across three environments: PointGoal1, PointButton1, and PointPush1.
The experimental results, reported in Figure \ref{fig:exp1}, 
demonstrate that our proposed SMPO method effectively utilizes the exploration cost 
to produce good average episode rewards 
while maintaining cumulative costs very close to the cost threshold. 
Notably, our proposed SMPO and the other two safe RL methods, 
PPO Lagrangian and CPQ, all approximately uphold the cost limit constraints 
indicated by the dashed black line,
while SMPO outperforms PPO Lagrangian and CPQ in terms of average episode reward across 
the three environments. 
While PPO yielded higher rewards than SMPO in PointGoal1 and PointButton1
and CPO produced higher rewards than SMPO in PointGoal1 and PointPush1,
both PPO and CPO largely violate the cost limit constraints in all the three environments,
thereby failing the goal of safe RL. 
These findings underscore the effectiveness of our proposed approach
for safe RL across different task environments.

\paragraph{Exploring Various Robots and Task Challenges}
%
\begin{figure}[t]
\centering
    {\includegraphics[width=0.32\textwidth]{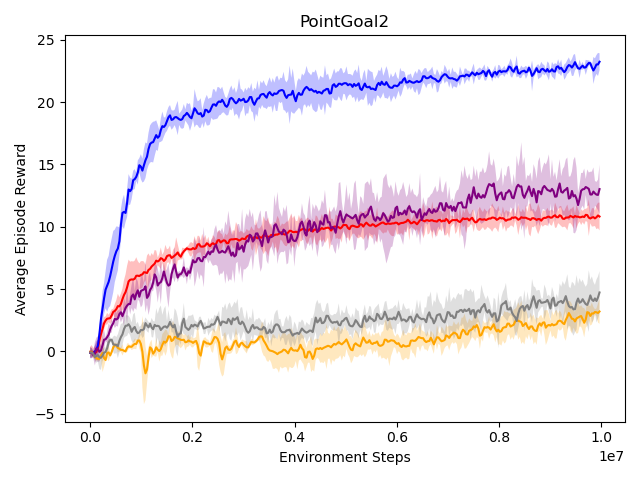}} 
	{\includegraphics[width=0.32\textwidth]{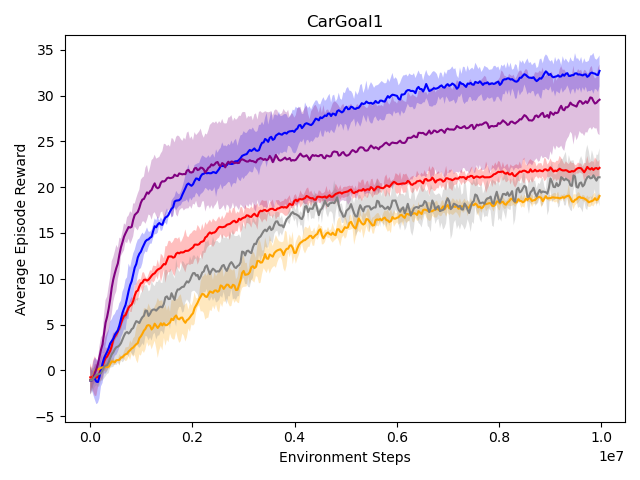}}
	{\includegraphics[width=0.32\textwidth]{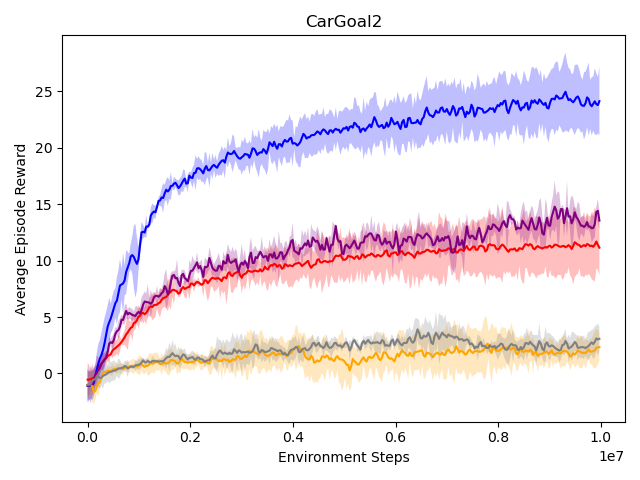}}
    {\includegraphics[width=0.32\textwidth]{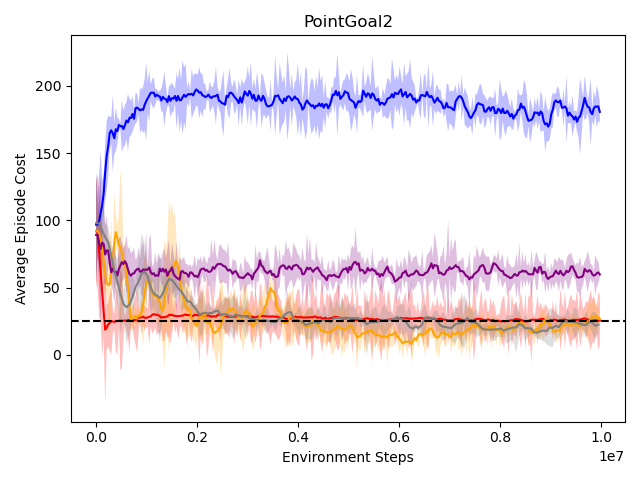}} 
	{\includegraphics[width=0.32\textwidth]{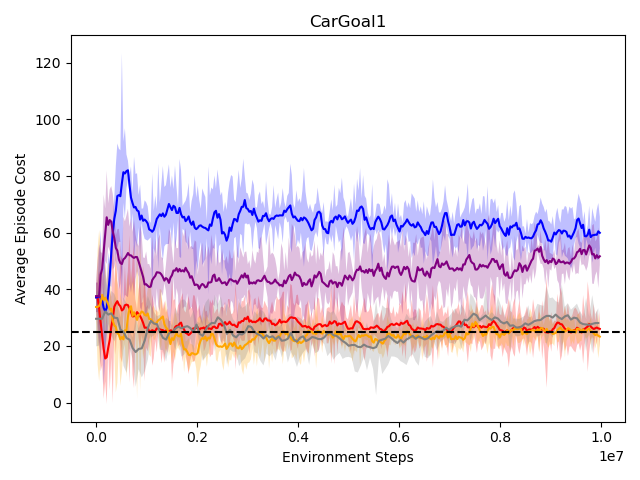}}
	{\includegraphics[width=0.32\textwidth]{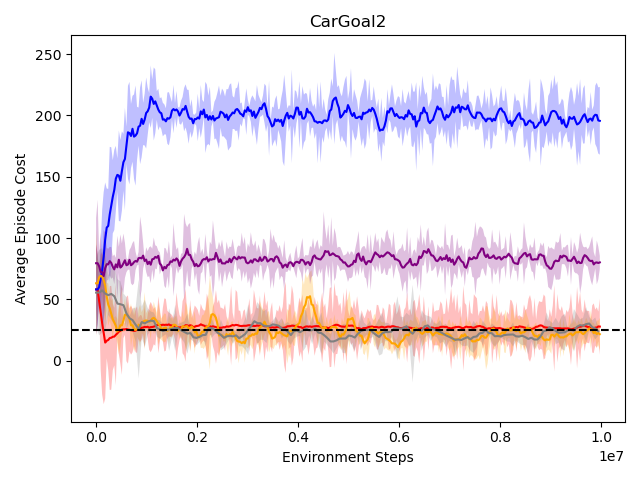}}
    {\includegraphics[width=0.5\textwidth]{figure/exp_legend.png}}
\caption{
Comparison results in terms of average episode reward/cost vs. environment steps 
on the Point robot with Goal2---a task with higher difficulty, 
and the Car robot for two tasks---Goal1 and Goal2. 	
The top row reports the results in terms of average episode rewards,
while the bottom row reports the corresponding average episode costs. 
The results are averaged over three runs, with the shadow indicating standard deviations and the dashed black line indicating the cost threshold.
}
\label{fig:exp2}
\end{figure}
We further evaluated the performance of our method across varying levels of task difficulty and different types of robots,
exploring diverse RL agents. 
Specifically, 
we conducted experiments
with the Point robot for the Goal2 task, 
which presents more hazards and obstacles in the environment exploration, 
and with the Car robot for the two tasks of varying difficulty levels: Goal1 and Goal2. 
The experimental results are presented in Figure \ref{fig:exp2}. 
Compared to CPO, our SMPO method demonstrates comparable performance in terms of average episode reward for the PointGoal2 and CarGoal2 tasks, while exhibiting significantly lower cumulative costs that are very close to the cost threshold. 
By contrast, PPO and CPO again significantly violated the safety limit constraints across all three tasks. 
Moreover, our SMPO method largely outperforms PPO Lagrangian and CPQ across 
all three tasks, while achieving similar levels of safety. 
This highlights the superior performance of our SMPO method, particularly on more challenging safe RL tasks. 
These results underscore the effectiveness of our proposed approach
for safe RL across different types of robots.
%

\subsection{Ablation Study}
\begin{figure}[t]
\centering
    {\includegraphics[width=0.24\textwidth]{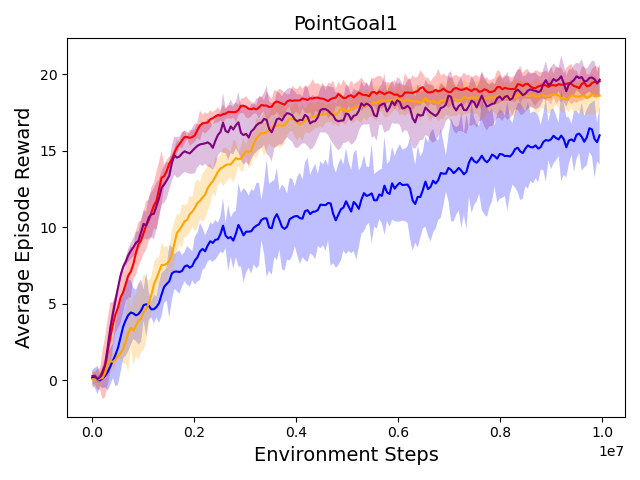}} 
	{\includegraphics[width=0.24\textwidth]{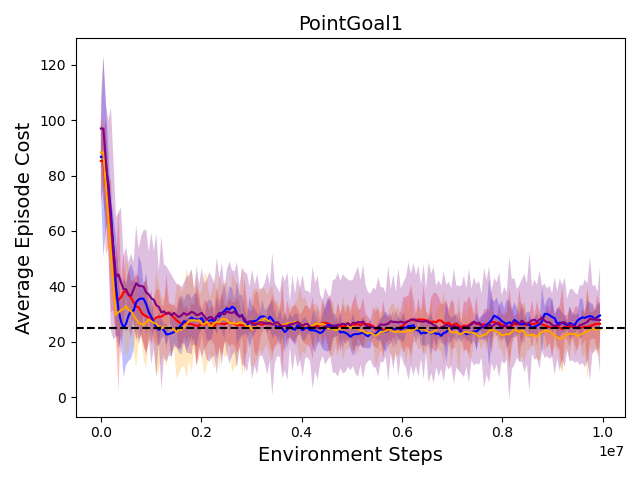}}
	{\includegraphics[width=0.24\textwidth]{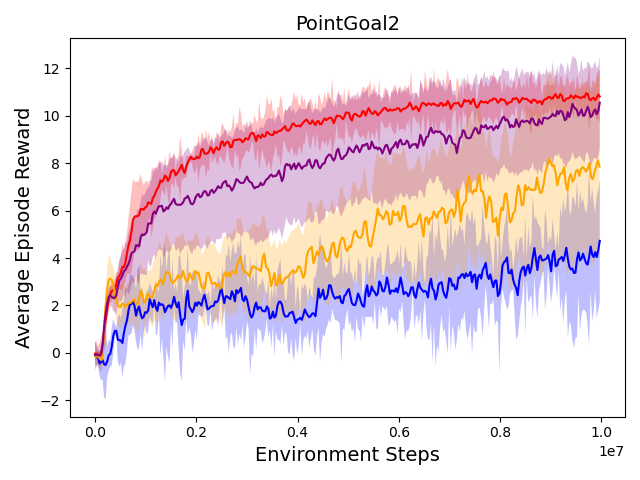}}
    {\includegraphics[width=0.24\textwidth]{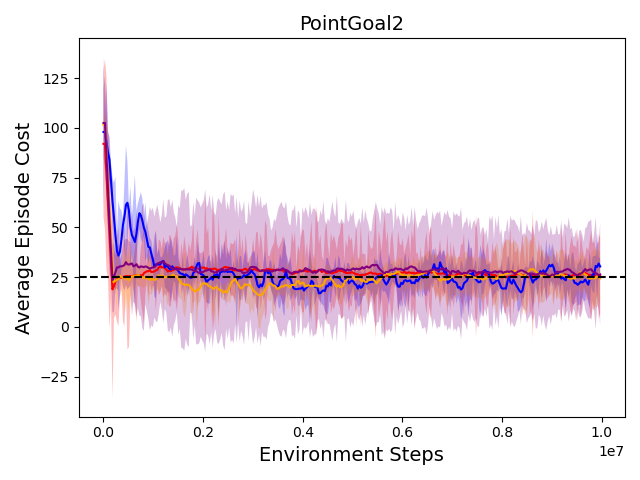}}\\
    {\includegraphics[width=0.75\textwidth]{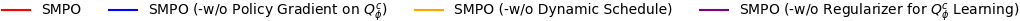}}
\caption{
Results of ablation study for the proposed SMPO method on PointGoal1 and PointGoal2. 
SMPO is compared with three ablation variants: 
``-w/o Policy Gradient on $Q_\phi^c$'', ``-w/o Dynamic Schedule'', and ``-w/o Regularizer for $Q_\phi^c$ Learning''.
	The results are averaged over three runs, with the shadow indicating standard deviations and the dashed black line indicating the cost threshold.
}
\label{fig:ablation}
\end{figure}
We conducted an ablation study on the SMPO method to validate the contribution 
of several components. 
In particular, we compared the full SMPO method with three ablation variants: 
(1) ``-w/o Policy Gradient on $Q^c_\phi$'', 
which drops the gradient with respect to the safety critic $Q^c_\phi$ in Eq.(\ref{eqa:gradient_policy}) 
for policy gradient update.
(2) ``-w/o Dynamic Schedule'', 
which drops the dynamic schedule on cost threshold and instead employs a fixed cost threshold $d$ 
throughout the entire training procedure.
(3) ``-w/o Regularizer for $Q_\phi^c$ Learning'', 
which drops the regularization term in Eq.(\ref{eqa:safety_critic})
by setting $\lambda=0$.

The results of the ablation study on PointGoal1 and PointGoal2
are presented in Figure \ref{fig:ablation}. 
As shown in the plots, the impact of these variants is mainly on the performance in terms of reward.  
The episode reward 
obtained by the variant ``-w/o Policy Gradient on $Q_\phi^c$'' 
increases substantially slower than that of the full SMPO method, 
with significantly higher variance during training 
in both environments. 
This validates the direct contribution of the safety critic $Q_\phi^c$ on policy gradient
through the weighting function for accelerating the learning of safe policy functions. 
The variant ``-w/o Dynamic Schedule'' 
exhibits much lower performance in terms of episode reward 
than the full SMPO method
during the early stage of training in the PointGoal1 environment.
In the PointGoal2 environment, this ablation variant demonstrates significantly lower episode rewards
and higher variances compared to SMPO, 
indicating that in certain circumstances 
the agent can fail to learn an effective policy without a good early learning period with relaxed cost limit. 
Compared to the full SMPO, the variant ``-w/o Regularizer for $Q_\phi^c$ Learning'' 
also exhibits noticeable decreases in episode rewards during a large portion of the training process,
but eventually reaches a similar performance level at the end of the training process. 
Moreover, it produces slower convergence and higher variance in terms of cumulative cost in the PointGoal1 environment, 
as well as significantly higher variances in terms of both episode reward and cost in the PointGoal2 environment. 
These observations underscore the importance of introducing regularization to stabilize the learning 
of the safety critic and its value to our safe RL approach. 
%

\subsection{Impact of Exponential Base}
\begin{figure}[t]
\centering
    {\includegraphics[width=0.24\textwidth]{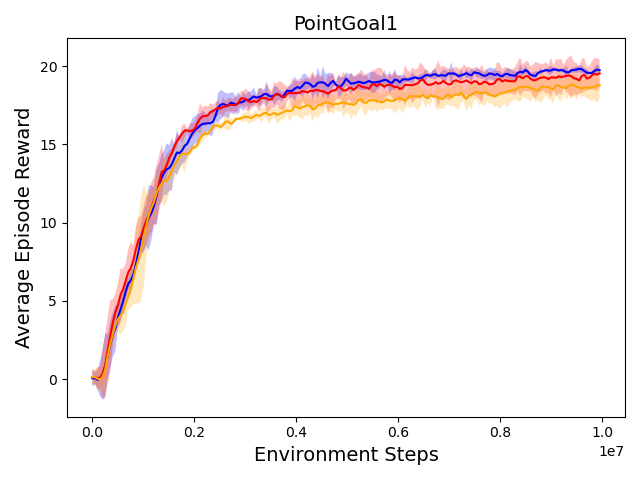}} 
	{\includegraphics[width=0.24\textwidth]{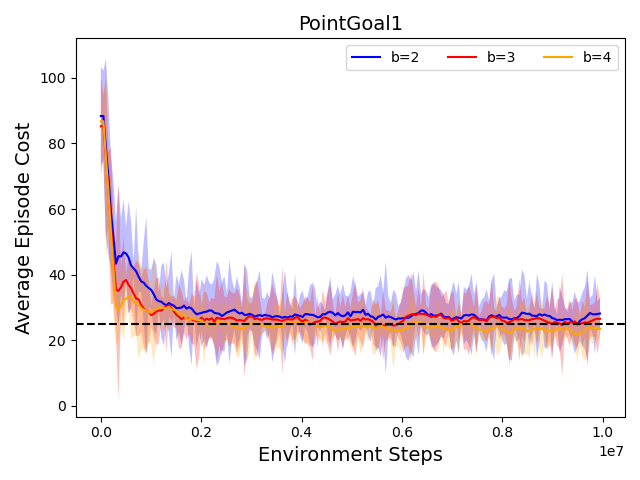}}
	{\includegraphics[width=0.24\textwidth]{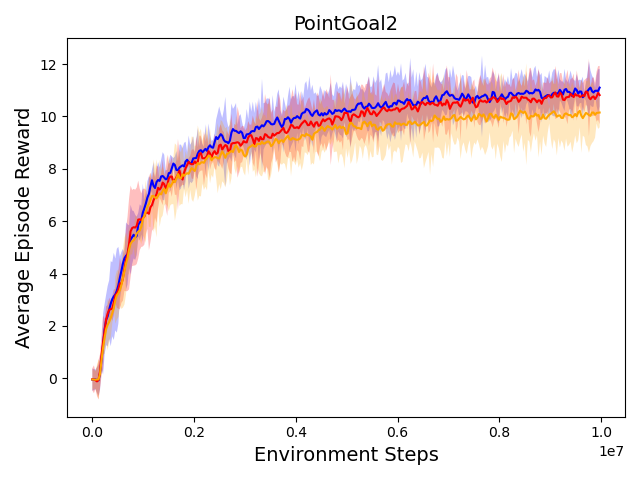}}
    {\includegraphics[width=0.24\textwidth]{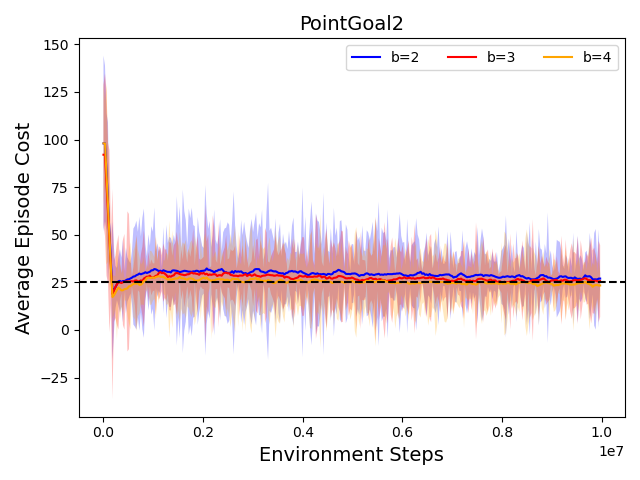}}\\
    {\includegraphics[width=0.25\textwidth]{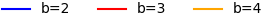}}
\caption{
	Results with different exponential base values, $b\in\{2,3,4\}$, for the cost-aware weighting function
	on the tasks of PointGoal1 and PointGoal2. 
	The results are averaged over three runs, with the shadow indicating standard deviations and the dashed black line indicating the cost threshold.
}
\label{fig:impact}
\end{figure}
We conducted an analysis on the impact of exponential base $b$ value used in the cost-aware weighting function 
on the performance of the SMPO algorithm. 
We tested different $b$ values
from the set of $\{2, 3, 4\}$,
with $b=3$ being the value employed in the main experiments. 
The results are depicted in Figure \ref{fig:impact}. 
It is evident that $b$ significantly influences 
the cost-aware weighting function,
thereby impacting the encoding of the cost limit constraints. 
A higher value of $b$ tends to better encode the cost limit constraints,
but yields slightly lower episode rewards than using a smaller $b$ value. 
The choice of exponential base $b=3$ indicates a good trade-off between episode cost and reward.

\section{Conclusion} 
Inspired by the growing safety consideration in reinforcement learning, 
we introduced a Safety Modulated Policy Optimization (SMPO) method, 
which provides a novel solution to safe RL by transforming
a constrained MDP problem to a standard unconstrained policy optimization problem.
The innovative form of reward modulation through a cost-aware weighting function
automatically ensures the safety constraint 
within standard policy optimization frameworks 
with the support of a safety critic. 
During RL exploration, the safety critic and the policy function can be
simultaneously learned within the SMPO framework to accelerate and enhance the safe learning process. 
Experimental results confirm that SMPO firmly adheres to safety constraints and 
outperforms state-of-the-art methods in terms of the overall performance-safety trade-off.

\bibliography{rlc2024}
\bibliographystyle{rlc2024}

\newpage
\appendix
\section{SMPO Algorithm}
\begin{algorithm}[th]
\caption{Safety Modulated Policy Optimization}
\label{alg:smpo}
\textbf{Input:} 
cost threshold $d$, base of weighting function $b$, epoch threshold for dynamic schedule $e_{\max}$ \\
\textbf{Initialize:}
safety critic $Q_\phi^c$, policy function $\pi_\theta$\\
\textbf{Procedure:}
\begin{algorithmic}[1]
\For{epoch $e_p=0,1,...,E_p$}
\State Compute the dynamically scheduled cost threshold $d'$ via Eq.(\ref{eq:dschedule}).
\State Set reply buffer $D=\emptyset$
\For{episode $e_s=0,1,...,E_s$}
\State Randomly initialize a start state $s_0$.
\For{environment step $t=0,1,...,T$}
    \State Sample action $a_t$ from policy: $a_t\sim\pi_\theta(\cdot|s_t)$.
    \State Collect transition $(s_t,a_t,r_t,c_t,s_{t+1})$ from the environment.
    \State Compute cumulative cost $\sum_{k=0}^{t-1} c_t$.
	\State Compute clipped future cost value $Q^c_{\phi,\text{clip}}(s_t,a_t)=\text{clip}(Q_\phi^{c}(s_t,a_t), 0, d)$.
    \State Add transition: 
	$D=D\cup\{(s_t,a_t,r_t,c_t,s_{t+1},\sum_{k=0}^{t-1} c_t+Q^c_{\phi,\text{clip}}(s_t,a_t))\}$.
\EndFor
\EndFor
\For{each gradient step}
	\State Collect transitions 
	$\{(s_t,a_t,r_t,c_t,s_{t+1},\sum_{k=0}^{t-1} c_t+Q^c_{\phi,\text{clip}})\}$ from replay buffer $D$.
    \State Compute safety modulated reward $\Mc(r_t)$ via Eq.(\ref{eq:weightfunc}) and Eq.(\ref{eqa:safety_modulated}).
    \State Update safety critic $Q_\phi^c$ by minimizing Eq. (\ref{eqa:safety_critic}).
    \State Compute gradient term of Eq. (\ref{eqa:gradient_policy}) by solving Eq. (\ref{eqa:gradient_critic}) and Eq. (\ref{eqa:gradient_reward}).
\State Update policy $\pi_\theta$ by applying policy gradient in Eq. (\ref{eqa:gradient_policy}).
\EndFor
\EndFor    
\end{algorithmic}
\end{algorithm}
Our SMPO algorithm is outlined in Algorithm \ref{alg:smpo}. 
At the beginning of each epoch, the dynamically scheduled cost threshold $d'$ 
is computed based on the current epoch number and will be used as the cost threshold $d$ in this epoch. 
The replay buffer $D$ is cleared at the start of each epoch. Subsequently, each episode within the epoch is initialized with a random start state $s_0$. 
At every environment step $t$ within each episode, an action $a_t$ is sampled from the policy function 
$\pi_\theta(\cdot|s_t)$, producing a transition $(s_t,a_t,r_t,c_t,s_{t+1})$ by interacting with the environment.
The cumulative cost $\sum_{k=0}^{t-1} c_t$ and the clipped safety critic value $Q^c_{\phi,\text{clip}}(s_t,a_t)$ 
are computed for each transition, 
which together will be used for computing the cost-aware weighting function value later. 
At each step, a transition ${(s_t,a_t,r_t,c_t,s_{t+1},\sum_{k=0}^{t-1} c_t+Q^c_{\phi,\text{clip}})}$ 
is added to the replay buffer $D$.
Upon completing the data collection for each epoch, 
we update both the safety critic $Q_\phi^c$ and the policy function $\pi_\theta$ 
using the sampled transitions from the replay buffer $D$ 
for each gradient step.


\end{document}